\documentclass[iicol, pdflatex]{sn-jnl}

\usepackage{graphicx}%
\usepackage{multirow}%
\usepackage{amsmath,amssymb,amsfonts}%
\usepackage{amsthm}%
\usepackage{mathrsfs}%
\usepackage[title]{appendix}%
\usepackage[dvipsnames]{xcolor}%
\usepackage{textcomp}%
\usepackage{manyfoot}%
\usepackage{booktabs}%
\usepackage{algorithm}%
\usepackage{algorithmicx}%
\usepackage{algpseudocode}%
\usepackage{listings}%
\usepackage{subcaption}%

\usepackage[backend=biber, sorting=none, style=numeric-comp]{biblatex}
\usepackage{makecell}
\usepackage{lipsum}
\usepackage{anyfontsize}
\usepackage[capitalize]{cleveref} 

\renewcommand{\cite}{\supercite}

\crefname{appendix}{Supplementary Note}{Supplementary Notes}
\Crefname{appendix}{Supplementary Note}{Supplementary Notes}
\AtBeginEnvironment{appendices}{\crefalias{section}{appendix}}

\setcellgapes{4pt} 

\usepackage{lineno}

\newcommand{\suppmaketitle}{\newpage\null%
    \hsize\textwidth\parindent0pt%
    \begin{center}
        {\LARGE \textbf{Supplementary Material}}\\[10pt]
        {\Titlefont An open-source tool for mapping war destruction at scale in Ukraine using Sentinel-1 time series\par} 
    \end{center}

    \global\punctcount\aucount%
    \vskip20pt%
    {
        \artauthors\par
        {
            \vskip7pt\addressfont\auaddress\par
        \removelastskip\vskip24pt%
        \ifnum\emailcnt>0\relax%
            \ifx\corrauthemail\@empty\else{\ifnum\aucount>1*\fi}%
           Corresponding author(s). E-mail(s): \corrauthemail\par\fi%
            \fi%
        }
    } 
    \vskip36pt 
    \removelastskip
    \noindent 
}

\begin{document}
\begin{refsection}[references.bib]

\title[Article Title]{An open-source tool for mapping war destruction at scale in Ukraine using Sentinel-1 time series}


\author*[1]{\fnm{Olivier} \sur{Dietrich}}\email{odietrich@ethz.ch}
\author[1]{\fnm{Torben} \sur{Peters}}
\author[2]{\fnm{Vivien} \sur{Sainte Fare Garnot}}
\author[3]{\fnm{Valerie} \sur{Sticher}}
\author[4]{\fnm{Thao} \sur{Ton-That Whelan}}
\author[1]{\fnm{Konrad} \sur{Schindler}}
\author[2]{\fnm{Jan Dirk} \sur{Wegner}}

\affil[1]{\orgname{Photogrammetry and Remote Sensing, ETH Zurich}, \city{Zurich}, \country{Switzerland}}
\affil[2]{\orgname{EcoVision Lab, Department of Mathematical Modeling and Machine Learning, University of Zurich}, \city{Zurich}, \country{Switzerland}}
\affil[3]{\orgname{Department of Political Science, University of Zurich}, \city{Zurich}, \country{Switzerland}}
\affil[4]{\orgname{International Committee of the Red Cross}, \city{Geneva}, \country{Switzerland}}



\abstract{Access to detailed war impact assessments is crucial for humanitarian organizations to assist affected populations effectively. However, maintaining a comprehensive understanding of the situation on the ground is challenging, especially in widespread and prolonged conflicts. Here we present a scalable method for estimating building damage resulting from armed conflicts. By training a machine learning model on Synthetic Aperture Radar image time series, we generate probabilistic damage estimates at the building level, leveraging existing damage assessments and open building footprints. To allow large-scale inference and ensure accessibility, we tie our method to run on Google Earth Engine. Users can adjust confidence intervals to suit their needs, enabling rapid and flexible assessments of war-related damage across large areas. We provide two publicly accessible dashboards: a \href{https://olidietrich.users.earthengine.app/view/ukraine-damage-explorer}{Ukraine Damage Explorer} to dynamically view our precomputed estimates, and a \href{https://olidietrich.users.earthengine.app/view/rapid-damage-assessment-sentinel1}{Rapid Damage Mapping Tool} to run our method and generate custom maps.}




\maketitle

\begin{figure*}[!htpb]
    \centering
    \includegraphics[width=.9\textwidth]{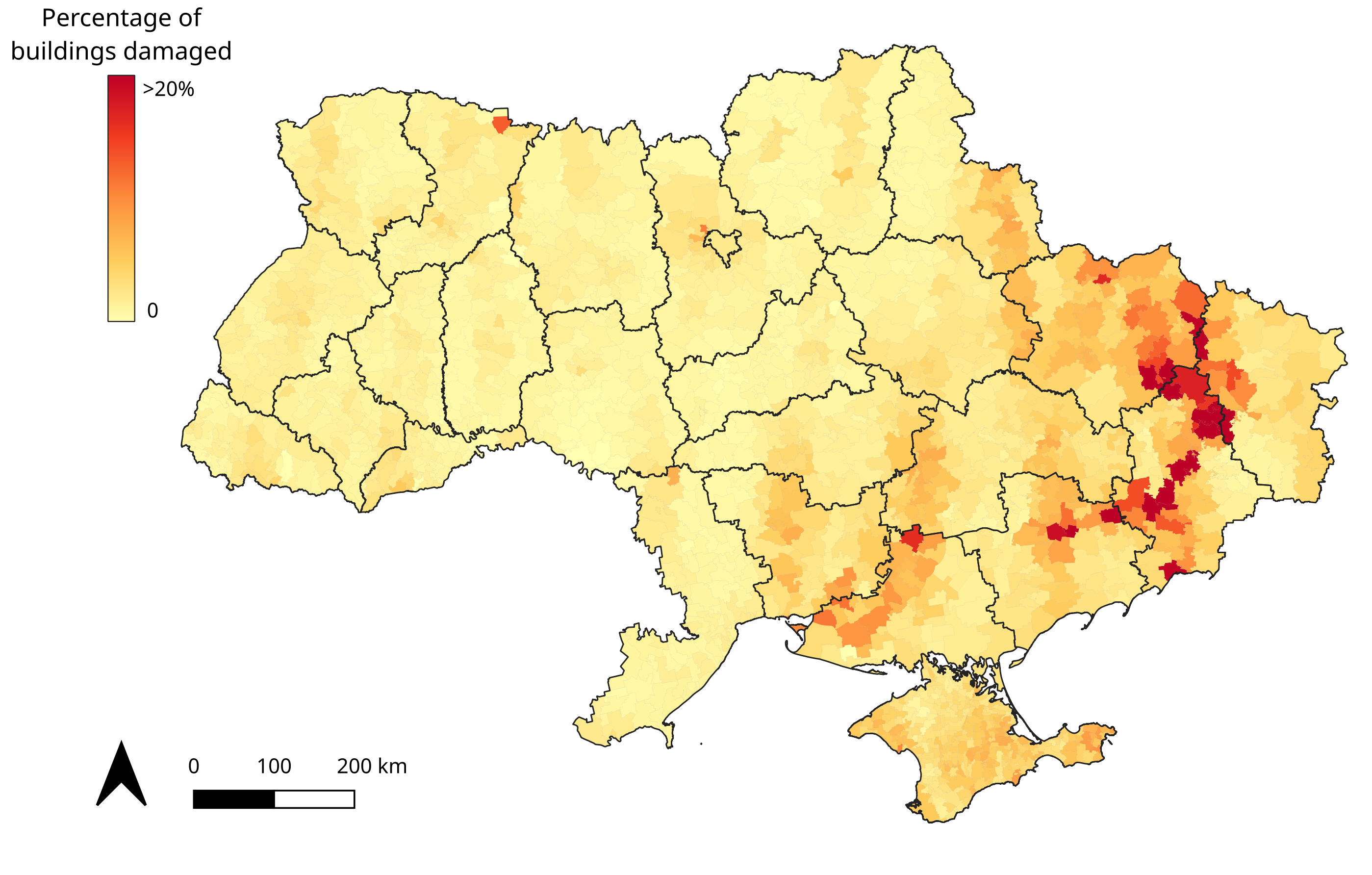}
    \caption{\textbf{Spatial distribution of building damage in Ukraine after two years of conflict.} Percentage of buildings likely damaged within the first two years of the war, aggregated by \textit{hromadas}. The predictions were thresholded at 0.655, and only buildings larger than 50$\,$m\textsuperscript{2} were considered.}
    \label{fig:admin3_stats}
\end{figure*}

\section*{Introduction}

The Russian invasion of Ukraine in February 2022 escalated a simmering conflict, which had previously seen unofficial Russian involvement limited to the country's eastern regions, into a full-scale war between the two countries. Russian troops crossed into Ukraine from multiple fronts and advanced through the country, reorienting their geographical focus as Ukrainian forces mounted an unexpectedly fierce resistance. As the Russian troops advanced, they subjected numerous cities to heavy shelling and destroyed critical infrastructure. Millions of Ukrainians fled their homes as damage and destruction spread throughout the country. Two years into the conflict, the war has caused hundreds of thousands of casualties and inflicted billions of dollars in damages to Ukraine's infrastructure \cite{kse_russia_will_pay, WorldBank2023}.

The fast-paced early stages of the Russia-Ukraine war serve as a stark reminder of the challenges humanitarian organizations encounter when monitoring ongoing armed conflicts in real-time. Even in less intense conflicts, like the war in Eastern Ukraine before the full-scale Russian invasion, violence can flare up unexpectedly, and minor incidents may trigger escalatory dynamics. Thus, regardless of a conflict's intensity, it is essential for humanitarian organizations to have a complete and up-to-date understanding of the situation on the ground to effectively support those most affected by the conflict. However, the task of maintaining such an overview is challenging, especially with limited resources. The difficulty is compounded when conflicts affect large geographic areas, persist over long periods of time, or occur in regions that are inaccessible due to security concerns.\\

To overcome these challenges, organizations have increasingly relied on satellite imagery to supplement on-the-ground monitoring. Although satellite images lack contextual details and a narrative dimension, they provide snapshots of the conditions on the ground, making it possible to observe how the situation evolves and how destruction spreads. Furthermore, satellites offer global coverage and automatic data acquisition, which is essential for monitoring large, remote, or inaccessible areas. The primary technique to perform damage assessments from satellite imagery is manual analysis of very high-resolution (VHR) optical images. As it involves the visual comparison of expensive commercial imagery, that method is labor-intensive and costly. Despite its accuracy and reliability, it depends on the availability of VHR data and does not scale to conflict zones that are large and that require long-term monitoring.

Nowadays, the analysis of satellite imagery with machine learning offers opportunities to automate parts of the labor-intensive remote monitoring process. Multiple studies have demonstrated the efficiency and effectiveness of this approach for assessing building damage. The most accurate methods currently involve applying deep neural networks to VHR optical data~\cite{Gueguen2015, Xu2019, Lee2020, Zheng2021, Gholami2022, Kaur2023}, often leveraging large dedicated datasets~\cite{xBD}. These approaches primarily focus on natural disasters, which typically require a one-time analysis of relatively localized, clustered damages. In contrast, armed conflicts can last for months, years, or even decades, resulting in spatially dispersed and gradually increasing destruction. Consequently, they require continuous monitoring over extended periods, so that automated change detection can on the one hand bring larger benefits, but on the other hand, is more challenging ~\cite{Sticher2024}.

Recent work has begun to apply research specifically to the task of assessing building destruction induced by armed conflicts. In a seminal study, \citeauthor{Mueller2021} trained a CNN with VHR optical imagery and labels from UNOSAT~\cite{unosat} to assess war-induced building destruction in Syria, leveraging the persistent nature of war damage over time due to the absence of reconstruction efforts amidst ongoing fighting~\cite{Mueller2021}. However, accessing VHR optical imagery for conflict zones is costly, as commercial sources typically do not provide free access to their database in these situations~\cite{Bennett2022}, unlike after natural disasters~\cite{MAXAROpen, PlanetDisaster}. Despite being an active field of research~\cite{Nabiee2022, Singh2023}, relying on repeated, large-area coverage with VHR images to monitor armed conflicts at scale is not realistic for most actors. Very recently, \citeauthor{Hou2024} proposed a deep learning-based approach that can handle both VHR and moderate-resolution optical imagery~\cite{Hou2024}.

Beyond optical images, Synthetic Aperture Radar (SAR) imagery offers a promising alternative. SAR is an active sensor system that illuminates the Earth's surface with microwave pulses and captures the backscattered signals. Unlike its optical counterpart, SAR can operate at any time of day (night, respectively) and is largely unaffected by clouds. SAR has a well-established history in damage assessment for both natural and anthropogenic events~\cite{Plank2014, Ge2020}. Broadly, SAR-based change detection can be divided into incoherent and coherent methods. Incoherent approaches rely only on backscatter amplitude, typically comparing intensity or correlation differences between pre- and post-event images~\cite{Aoki1998, Matsuoka2004, Matsuoka2005, Liu2013, Uprety2013}. Coherent methods, on the other hand, use both amplitude and phase information through interferometry (InSAR), generally resulting in more accurate results~\cite{Matsuoka2000, Fielding2005, Gamba2007}. Additionally, SAR's sub-meter wavelength allows coherent techniques to detect changes far smaller than the pixel resolution, particularly when using permanent scatterers, which maintain coherence over long time series~\cite{Ferretti2001}. Coherent change detection is a standard approach in the InSAR community to assess building damage after a disaster rapidly~\cite{Yun2015, Watanabe2016, Washaya2018, Tay2020}, c.f.\ the popular (and patented) \textit{Damage Proxy Maps}. Recent works have combined machine learning and coherence data to produce more accurate maps, either with handcrafted features~\cite{Mastro2022, Akhmadiya2022} or by using recurrent neural networks trained on coherence time series~\cite{Stephenson2022, Yang2024}. On the downside, generating high-quality coherence maps is complex and involves a loss of resolution. In this work, we focus on large-scale deployment and accessibility and opted for developing our approach on the cloud-based platform Google Earth Engine (GEE)~\cite{Gorelick2017}. Since GEE does not support the Sentinel-1 SLC product~\cite{sentinel_1} necessary for phase-based methods, our approach is purely amplitude-based.

A few studies have already explored satellite-based damage assessment for the Ukraine War, but most focused on specific areas or events. \citeauthor{Aimaiti2022} achieved 58\% accuracy on UNOSAT labels in the Kyiv region based on handcrafted features from Sentinel-1 and Sentinel-2~\cite{Aimaiti2022}. \citeauthor{Huang2023} used optical and coherence SAR time series to identify burnt areas, respectively urban destruction in Mariupol, reaching an overall accuracy of $\approx$60\% across four classes relative to labels retrieved from social media images~\cite{Huang2023}. \citeauthor{Tavakkoliestahbanati2024} leverage multi-temporal InSAR to monitor deformation from a single event, the Kakhovka dam collapse~\cite{Tavakkoliestahbanati2024}. \citeauthor{Ballinger2024}'s method is arguably the one most directly comparable to ours. It relies on a pixel-wise statistical $t$-test on Sentinel-1 time series, implemented directly in GEE. UNOSAT labels from various locations were then used to calibrate the decision threshold~\cite{Ballinger2024}. Lastly, \citeauthor{Scher2023} have generated nationwide coherence maps and performed coherent change detection against a reference pre-invasion image. At the time of writing their maps have not yet been made public~\cite{Scher2023}.\\

In this work, we introduce an easy-to-use, open-source war impact mapping tool based purely on SAR amplitude data and demonstrate its effectiveness through a comprehensive building damage assessment across Ukraine. Our solution leverages existing, point-wise damage maps from UNOSAT~\cite{unosat} and public, open SAR imagery from the Sentinel-1 mission~\cite{sentinel_1}. Harnessing the parallel computing capabilities of GEE~\cite{Gorelick2017}, we train a model to estimate the likelihood of war-related changes from paired time series of Sentinel-1 backscatter. We choose a fixed one-year interval in 2020 as our reference period and iteratively generate damage likelihood maps for 3-month periods ranging from February 2021 up to February 2024. Our model, based on the Random Forest algorithm, allows for straightforward deployment on GEE and outputs damage maps with 10$\times$10$\,$m grid spacing, the original ground sampling distance (GSD) of the Sentinel-1 GRD product. In a second processing step, we intersect the per-pixel maps with publicly available building footprints from Overture Maps~\cite{OvertureMaps} to generate a damage estimate per building. Our tool produces maps of war damage at a very low cost and enables continuous country-scale monitoring. These maps can serve both as a basis for large-scale, aggregated assessments and as guidance to focus manual verification efforts for individual settlements or buildings.
We offer two public online dashboards: In the
\href{https://olidietrich.users.earthengine.app/view/ukraine-damage-explorer}{first one}, users can dynamically view and explore the pre-computed damage maps described in the following. The \href{https://olidietrich.users.earthengine.app/view/rapid-damage-assessment-sentinel1}{second one} allows users to run our method themselves for their desired locations and time periods and create custom maps.

The key advantages of our approach are its scalability, accessibility, and ability to transfer to different geographic contexts. Using commercial VHR imagery for regular screening across entire conflict zones is not viable for humanitarian organizations~\cite{Sticher2023}, because it would be too resource-intensive, in terms of both direct data cost as well as infrastructure for data download, storage, and compute. By utilizing moderate-resolution satellite images that are free of charge and acquired with short, regular revisits, we enable a more sustainable monitoring approach. The use of SAR imagery has the additional advantage that the same method can be expected to work well in other conflict contexts, including those with different architectural patterns and/or with even more frequent cloud cover, as it is purely geometrical and does not rely on any location-specific a-priori assumptions. The proposed method is also fast, reproducible, and easy to adapt, even for users who are not experts in remote sensing or image analysis. We believe that our work could be a valuable and accessible resource and could be used to screen large conflict areas beyond our specific application case.

\section*{Results}

\subsection*{Evaluation}
\label{sec:eval}

\begin{figure*}[!htpb]
    \centering
    \includegraphics[width=.95\textwidth]{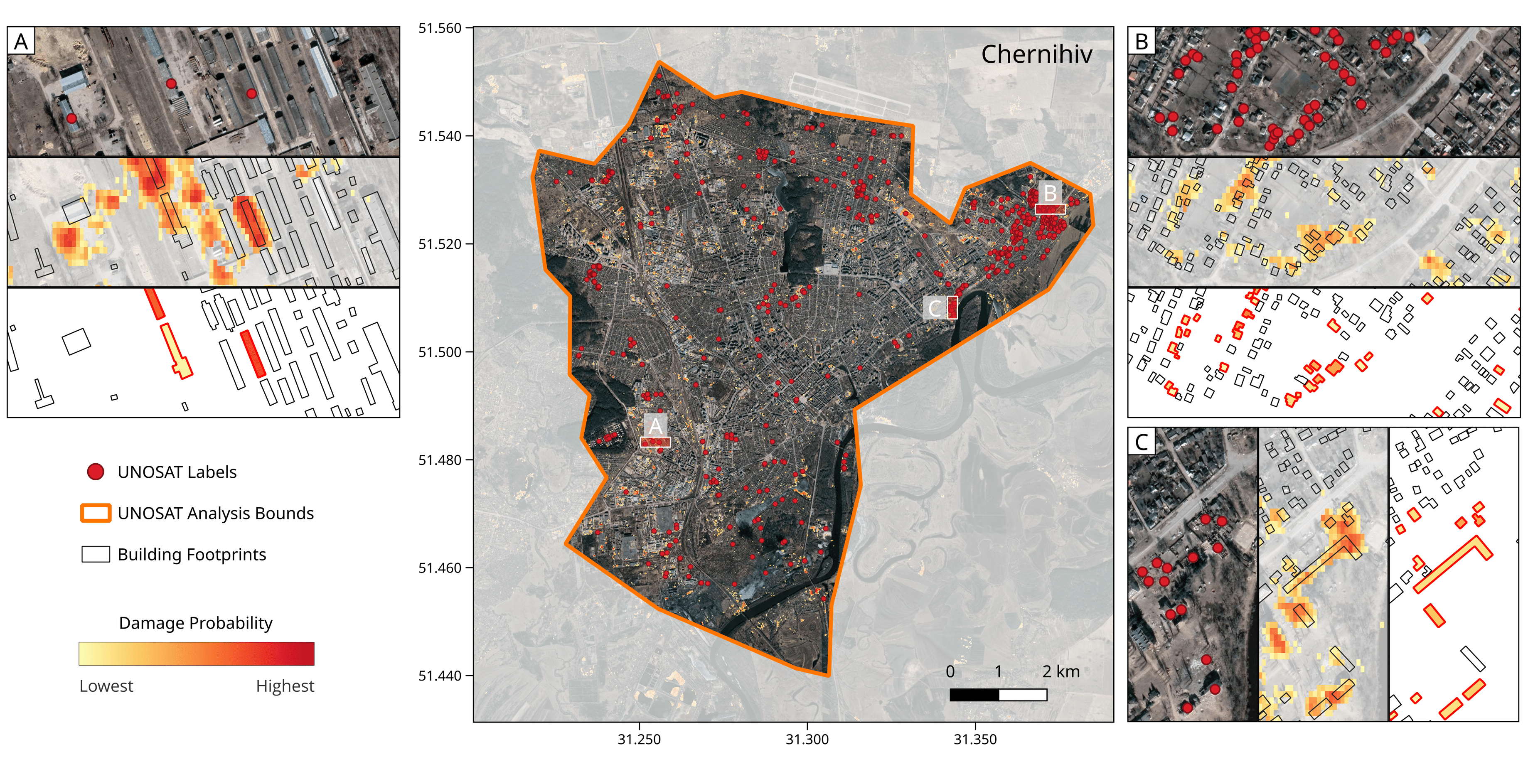}
    \caption{\textbf{Damage estimates and UNOSAT labels for Chernihiv.} Building damage estimate for Chernihiv thresholded with the standard confidence threshold of 0.655, aggregated over the first two years of conflict. For clarity, we only present the damage heatmap in the main figure, while three zoomed insets show building-level predictions. Red dots represent UNOSAT annotations indicating buildings marked as either \textit{destroyed} or \textit{severely damaged}. The VHR satellite layer is displayed solely for visualization, all results are derived from 10$\,$m-resolution Sentinel-1 images. Readers are encouraged to explore the maps on the interactive \href{https://olidietrich.users.earthengine.app/view/ukraine-damage-explorer}{dashboard}. Sources: Google Earth/Maxar Technologies, Overture Maps building footprints, UNITAR/UNOSAT damage annotations.}
    \label{fig:chernihiv_results}
\end{figure*}

\begin{table*}[!htpb]
\centering
\small
\begin{tabular}{lcccccc}
\toprule
\thead{Label}     & \thead{Precision} & \thead{Recall} & \thead{F\textsubscript{1}} & \thead{Accuracy} & \thead{AUC}\\
\midrule
Damaged     & 67.1 (90.3) & 84.6 (46.5) & 74.9 (61.4) & \multirow{2}{*}{80.3 (79.7)} & \multirow{2}{*}{81.3 (71.9)}\\
Undamaged   & 90.5 (77.4)  & 78.0 (97.3) & 83.8 (86.2) & \\
\bottomrule
\\
\end{tabular}
\caption{\textbf{Evaluation metrics.} Quantitative detection performance relative to UNOSAT labels (at previously unseen locations), using a confidence threshold of 0.5. Values in parentheses indicate performance at the threshold of 0.655, reaching a target precision of 90\%. The \emph{undamaged} class is evaluated using the same locations, but acquisition windows before the full-scale invasion. All metrics are given as percentages.}
\label{tab:pixel-wise-eval}
\end{table*}

We first evaluate our model's predictions against additional damage assessments excluded from the training set (see \cref{fig:unosat_data_overview} for the geographic distribution of our test set). The UNOSAT-labeled locations are compared to the corresponding raw pixel predictions of the model. True positive and false negative rates are counted in the 3-month periods from 2022 (after the invasion), and true negative and false positive rates are counted in the periods from 2021 (before the invasion). To guard against potential shifts between the point labels and the satellite tiles, e.g., due to high incidence angles or inaccurate annotation, we compare to the maximum predicted damage probability in a 3$\times$3 pixel window centered at the reference label. We report our results in \cref{tab:pixel-wise-eval}. We obtain an F1-score of 74.9\% for the damaged class and an AUC of 81.3\%. Using the same evaluation protocol, we compare our result to the pixel-wise $t$-test (PWTT) method~\cite{Ballinger2024}. To our knowledge, this is the only method directly comparable to ours, since others either require coherence maps or assume that the precise date of the destruction event is known, which is not the case for our UNOSAT labels. Our method consistently outperforms that baseline, see \cref{tab:comp_ttest}. For details about the comparison see \cref{app:ttest}.

\begin{table}[!htpb]
\setlength{\tabcolsep}{8pt} 
\renewcommand{\arraystretch}{1.2} 
\centering
\small
\begin{tabular}{lcccccc}
\toprule
\thead{Method} & \thead{Recall} & \thead{F\textsubscript{1}} & \thead{AUC}\\
\midrule
PWTT\cite{Ballinger2024} & 76.3 & 68.4 & 75.7\\
Ours & 84.6 & 74.9 & 81.3\\
\bottomrule
\\
\end{tabular}
\caption{\textbf{Comparison with PWTT.} Comparison of our method with the PWTT~\cite{Ballinger2024}, with the latter calibrated for Ukraine. Metrics were computed on our test set, using the same settings for both approaches. All metrics are given in percentages.}
\label{tab:comp_ttest}
\end{table}

The evaluation of held-out validation data also helps establish an appropriate threshold for distinguishing between damaged and undamaged pixels. Indeed, while the model has been trained on a balanced dataset, the actual distribution of damaged versus undamaged pixels is highly skewed, and the prevalence of undamaged pixels in the test set means that thresholding at damage probability 0.5 will cause a considerable number of false alarms. To address this, we set a target precision of 90\%, aligning with the reality of operational deployment, where it is crucial to keep the false positive rate at a manageable level. With this choice, we find an optimal threshold of 0.655. For details about the threshold optimization see \cref{app:threshold}. With that value, the recall reaches 46.5\%. We point out that there is an inherent trade-off between missed detections and false alarms, and the right threshold depends on the application task. Our dashboard allows the user to adjust it with a slider and explore how different thresholds affect the estimates. Results with the confidence threshold of 0.655 are also reported in \cref{tab:pixel-wise-eval}.

\begin{figure*}[!htpb]
    \centering
    \includegraphics[width=.95\textwidth]{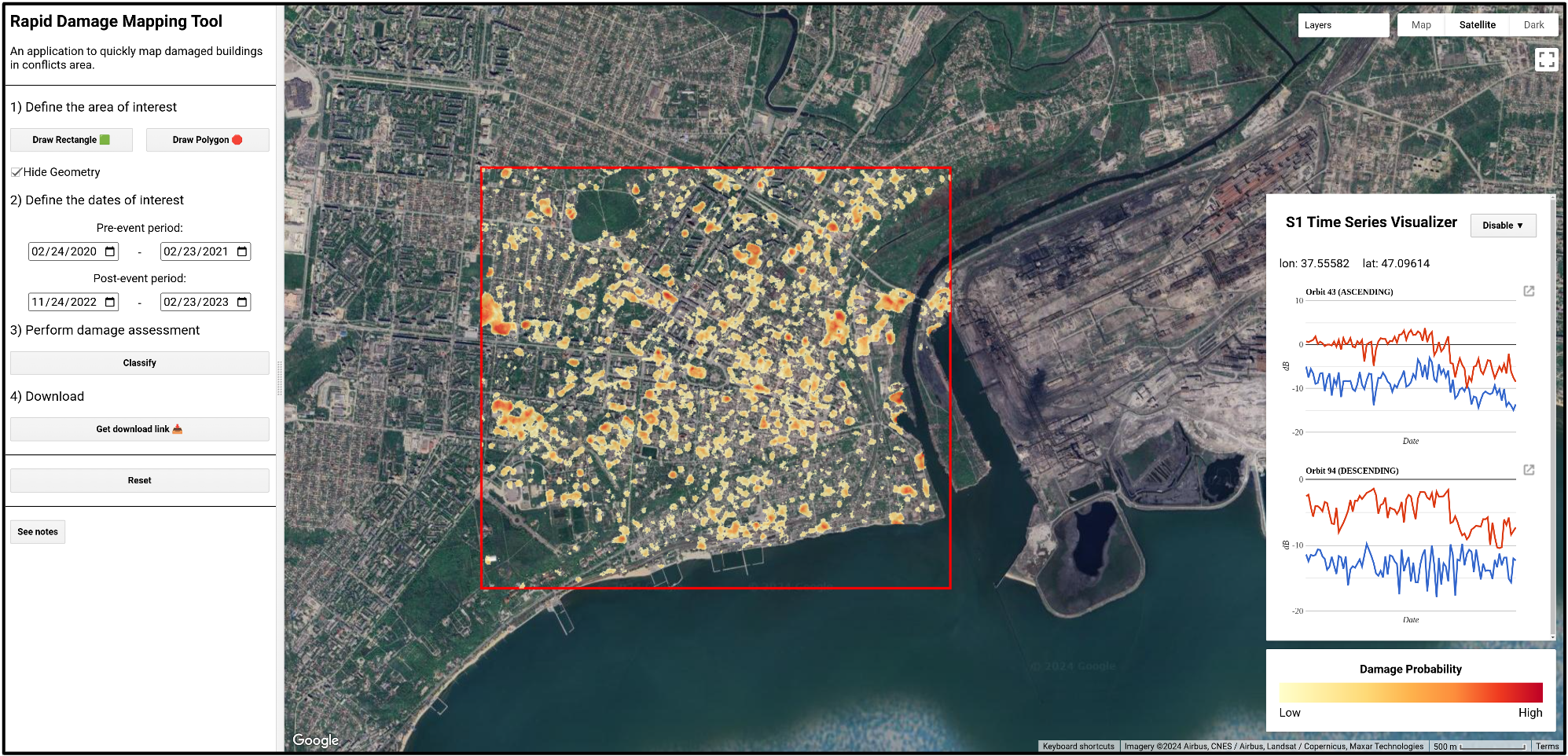}
    \caption{\textbf{The Rapid Damage Mapping Tool.} Screenshot from the \href{https://olidietrich.users.earthengine.app/view/rapid-damage-assessment-sentinel1}{dashboard interface}, displaying damage estimates for a region in Mariupol, alongside the Sentinel-1 time series visualizer.}
    \label{fig:webapp}
\end{figure*}

\subsection*{Close-Ups}
We use Chernihiv, one of the cities in our test set, to qualitatively illustrate the performance and limitations of our approach. \cref{fig:chernihiv_results} shows both heatmaps of damage probability and per-building maps after post-processing with building footprints. Inset A highlights one limitation of using building footprints, as the leftmost damage label has been correctly predicted in the heatmap but does not have a corresponding footprint. Inset B illustrates how the model can detect correct patterns of destruction, even for small structures. However, some labels (e.g., center-top) are completely missed, likely because the affected building volume was too small to be visible at Sentinel-1 resolution. Inset C shows both how multiple UNOSAT labels can fall within a single footprint. In the main figure, outside the boundaries of UNOSAT analysis, we observe agricultural fields wrongly identified as damaged in the heatmap. This is expected, since our model has not been trained for patterns that occur outside of settlement areas, and illustrates the need to post-process the raw heatmaps with settlement masks or building footprints.

\subsection*{Web-based Tool}
We have developed a web application based on Google Earth Engine (GEE) to facilitate rapid reproduction of our results and enable the generation of new maps without any local software installation. The application features a user-friendly dashboard (\cref{fig:webapp}) that provides the functionality to perform damage assessment for a user-specified region of interest and using user-defined pre- and post-event observation periods. The predicted maps can also be exported as raster files directly from the interface (subject to GEE restrictions on download size). Furthermore, the dashboard includes a visualization tool that enables pixel-by-pixel examination of Sentinel-1 time series data at any desired location.


\subsection*{Country-wide Analysis}
To illustrate the scalability of our approach, we have run damage assessments over the entire country. Our analysis indicates that over 400,000 buildings in Ukraine, or approximately 2.7\% of all buildings, have likely sustained damage during the first two years of conflict. This estimate only accounts for buildings larger than 50$\,$m\textsuperscript{2}. \cref{fig:admin3_stats} shows the percentage of buildings likely damaged within the first two years of the war, aggregated by \textit{hromadas}, the finest administrative division in Ukraine covering the entire territory. For an analysis of how damage has evolved over time, see \cref{app:temporal_evolution}. For the most impacted \textit{hromadas} with over 10,000 buildings---Bakhmutska, Marinska, and Popasnianska---damage rates reach 42.6\%, 29.0\%, and 25.0\%, respectively. We observe that this pattern of destruction correlates with the course of the war, the eastern side of the country having been a battlefield for months, while the western part has been relatively spared. For a subset of $\approx$1.8M buildings in our dataset, we were able to retrieve meta-data about the building function (such as "residential" or "medical") from the OpenStreetMap catalog. \cref{tab:damage_per_class} summarizes the damaged fractions for different building functions. In this preliminary analysis, we did not find any country-wide patterns that would suggest that certain building types are more prone to war damage than others. Note a more detailed study is needed to confirm or reverse this finding, as biases in the definition and distribution of the available OpenStreetMap building function labels could influence the result.

\begin{table}[htpb]
\centering
\begin{tabular}{lrr}
\toprule
\thead{OSM Class} & \thead{\# Damaged} & \thead{\% Total} \\ 
\midrule
Residential       & 57,089 & 3.6 \\
Industrial        & 3,177  & 5.0 \\
Outbuilding       & 599    & 1.6 \\
Commercial        & 1,309  & 3.9 \\
Education         & 658    & 3.5 \\
Agricultural      & 775    & 4.1 \\
Service           & 284    & 3.6 \\
Religious         & 177    & 2.1 \\
Medical           & 126    & 3.6 \\
Civic             & 117    & 4.6 \\
Transportation    & 106    & 4.9 \\
Entertainment     & 8      & 2.6 \\
Military          & 6      & 4.1 \\
\bottomrule
\end{tabular}
\caption{Number of buildings damaged per OSM class across Ukraine. The footprints with OSM class information represent a subset of 11.9\% of all building footprints analyzed.}
\label{tab:damage_per_class}
\end{table}


\section*{Discussion}

Knowing where and when war-related building damage has occurred is essential for humanitarian organizations to assist populations affected by armed conflicts. The open-access remote monitoring tool that we introduce in this article allows users to rapidly gain an overview of war-related impact. It can also supplement interactive damage mapping and serve as a pre-filtering step to guide reliable, but resource-intensive photo-interpretation by human operators. We have demonstrated the effectiveness of our tool by conducting a comprehensive building damage assessment across Ukraine, showing how war-related destruction has spread throughout the country. The retrieved distribution of damaged buildings broadly reflects the frontlines of the conflict, with the most severe destruction concentrated in areas that experienced prolonged and intense fighting.

While our method has so far been tested only in Ukraine, we plan to extend its application to other regions. Given the characteristics of both the Sentinel-1 SAR data and the Random Forest classifier, we anticipate that the methodology will adapt well to new areas, as long as we have adequate data for fine-tuning. However, the final performance will ultimately be influenced by factors such as vegetation cover and building density.

\subsection*{Limitations}

While our tool for screening conflict-related building damage proved effective, it is important to acknowledge its limitations and interpret its outputs accordingly. In this context, we again highlight that the purpose of our method is to complement existing damage mapping solutions with a scalable screening component, not to replace human photo-interpretation or ground-based damage reports.

\textbf{Classification Threshold:} One key limitation is the sensitivity of the quantitative estimates to the selected confidence threshold. As a pragmatic solution, our tool provides a slider that lets the user adjust the threshold according to the precision/recall trade-off needed for a specific use case. For example, a higher threshold may be appropriate to gather spatially coarse aggregate statistics about destruction patterns as in \ref{fig:admin3_stats}, whereas a lower threshold might better support the search for individual, hitherto overlooked damages.

\textbf{Google Earth Engine:} Our study focuses on large-scale assessment and accessibility, leveraging GEE’s computational capabilities. However, this reliance also imposes significant constraints. First, it restricts the choice of machine learning methods, as only a limited set of classification algorithms are available at the time of writing. Second, while Sentinel-1 data processed by GEE is orthorectified, it is not radiometrically terrain-flattened, which can introduce distortions in areas with pronounced slopes. Although most cities in Ukraine are situated on relatively flat terrain, our method may be less reliable in towns and cities located in mountainous regions. Additionally, as previously mentioned, GEE currently does not support the ingestion of complex SAR data, preventing the use of phase coherence for change detection. Lastly, while the proposed method is computationally efficient in principle, the web tool may take several minutes to generate results due to GEE’s technical overhead. As an alternative, users can download the relevant image tiles and process them locally using the provided code. In the future, we hope to further accelerate the online tool.

\textbf{Dependence on External Data:} Our model has specifically been trained on building data and therefore presupposes that settlement or building boundaries are available to mask out areas without human settlements (e.g., agriculture, forest, etc.), which would otherwise cause many false positives. We found that it is sufficient to utilize standard, global land cover maps~\cite{Brown2022}, settlement layers~\cite{gshl}, or building footprints. Still, the reliance on such external sources is a potential source of error (see \cref{app:footprints_limitations} for possible issues specific to building footprints).

\textbf{Choice of Reference Period:} To have a common pre-event reference for both negative and positive samples, we used the year 2020 as a fixed reference period throughout this work. This means, however, that the temporal gap between the pre-event and post-event periods can be rather large; giving rise to misclassifications due to surface changes other than war damage, e.g., construction. Ideally, a future version of the tool would integrate context knowledge about the conflict to minimize the temporal gap between the pre- and post-event observation periods.

\textbf{Reference Data:} Our supervised learning approach necessitates a sufficient volume of reference data. Despite existing calls for large-scale datasets and open data programs dedicated to armed conflicts~\cite{Bennett2022}, such resources remain limited. Data availability varies considerably between different contexts: for some conflicts, e.g.\ the wars in Ukraine and Gaza, quite a lot of damage data is available; whereas there may be very little information about building damages in conflicts that receive less public attention. It is precisely in these under-reported conflicts that automated analysis could have the biggest impact. By offering an accessible open-access tool based on public satellite data, we hope to take one step toward a global monitoring solution. But to get there it will likely be necessary to collect at least some training data that represent the building patterns and imaging conditions of regions with under-reported conflicts.


\subsection*{Policy Implications}

The combination of free Earth observation data and machine learning offers a promising route toward automated war damage monitoring, complementing interactive mapping and on-the-ground surveys.

But there is inherent uncertainty in predicting damage probabilities from SAR image patterns at 10$\,$m GSD, and users should understand this and align their expectations with what such methods can deliver. Our approach is perhaps best suited to obtain a high-level overview of how damage spreads across time and space. Beyond scientific goals like tracing the development of the conflict, this should often be enough to understand where populations are affected. At the same time, our method cannot confirm damage to an individual building, meaning that it is not directly comparable to assessments based on VHR imagery, ground-based photography, witness reports, etc. Importantly, the automatic large-scale approach and reliable local evidence are complementary: satellite damage maps can identify regions that warrant the effort to gather detailed, reliable evidence.

Post-processing the model outputs with additional data sources, as demonstrated for Ukraine with building footprints, adds an important dimension of information. Future research could seek to assimilate further sources of information, for instance, news reports or social media. Despite the inherent biases of such sources \cite{Miller2022}, they contain a wealth of complementary information that could be harvested, adding contextual knowledge and narrative depth to maps derived purely from satellite images. For example, attacks on residential buildings carry different policy implications if these buildings are inhabited than if civilians have fled an area in anticipation of strikes. Of course, even when data sources are combined to enhance the information derived from satellite images, contextual knowledge offered by observers and humanitarian organizations remains crucial for decision-making.

Beyond immediate humanitarian action, our large-area war impact maps are potentially useful also to support other efforts to remedy the consequences of armed conflicts, such as planning, prioritization, and resource allocation during reconstruction efforts. In terms of scientific inquiry, wall-to-wall, spatially explicit data allows for fine-grained geospatial analysis of conflict dynamics, which at present relies almost entirely on events derived from news articles. Building damage information obtained through our approach can serve as an additional data source that is complementary to the more commonly used metric of battle-related fatalities.

\section*{Methods}
\begin{figure*}[!htbp]
    \centering
    \includegraphics[width=.8\textwidth]{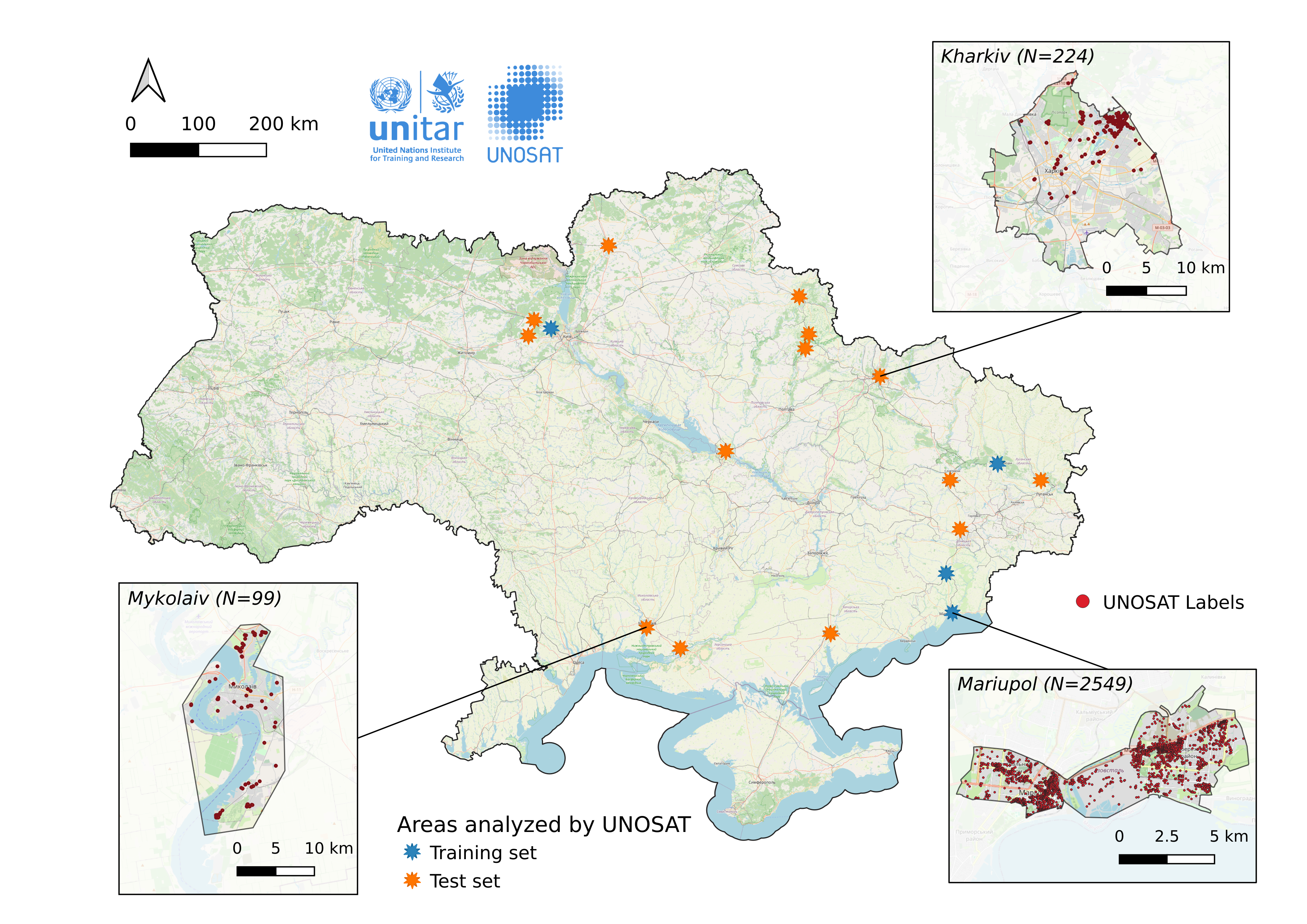}
    \caption{\textbf{Regions analyzed by UNOSAT.} Overview of the 18 regions assessed by UNOSAT in Ukraine. Blue and orange dots indicate cities used for training and testing, respectively. The three insets highlight the variation in damage density across different regions. Background map: OpenStreetMap.}
    \label{fig:unosat_data_overview}
\end{figure*}

\subsection*{Data}

All the data used in this work can be found freely available online.\\

\textbf{Reference Damage Assessments:} To the best of our knowledge, the only public, georeferenced datasets that provide building-level damage information for Ukraine are the UNOSAT maps~\cite{unosat}. They are based on commercial VHR imagery, where analysts manually compare pre- and post-event images using a standardized damage scale. To maximize the chances that these damages produce discernible changes in the SAR imagery, we only keep the two most severe damage levels, i.e., \textit{destroyed} or \textit{severely damaged}, and discard the others. We aggregate all available UNOSAT maps for Ukraine into a dataset consisting of 10,934 unique entries, distributed across 18 different areas of interest (AOIs) as depicted in \cref{fig:unosat_data_overview}. Importantly, these labels are attached to geo-coordinates, without any relation to concrete events. We, therefore, lack information on both the spatial extent and the precise date of the destruction, all that is known is that it happened between the pre- and post-event dates. For the present work, we further restrict this interval by assuming that no damage occurred before the onset of the invasion on February 24, 2022. We point out that this assumption may not strictly hold in the eastern regions, particularly Luhansk and Donetsk oblasts. A detailed summary of our dataset can be found in \cref{app:unosat_details}.\\

\textbf{SAR Image Time Series:} We use freely available SAR data from the Copernicus Sentinel-1 mission~\cite{sentinel_1}, consisting of two satellites equipped with C-band SAR instruments and moving in the same orbital plane, phased 180° apart. This setup originally allowed for global coverage with a revisit period below 6 days anywhere on Earth. Unfortunately, this period has doubled to up to 12 days since the technical failure of Sentinel-1B in December 2021. For our study, we only use data from Sentinel-1A to maintain consistent temporal resolution. We utilize the Ground Range Detected (GRD) product available on GEE, consisting of log-amplitudes for the VV and VH polarizations, resampled at 10$\,$m. Internally, GEE automatically preprocesses every tile with precise orbit correction, border and thermal noise removal, radiometric calibration, and terrain correction. We do not add any further preprocessing.\\

\textbf{Building Footprints:} We leverage the building layer from the Overture Maps Foundation's Open Map Data~\cite{OvertureMaps}, which combines building footprints from diverse sources. In Ukraine, this equates to a total of 26.5M buildings, with 24.7\% coming from OpenStreetMap's crowd-sourced database~\cite{OpenStreetMap} and the remaining 20M from Microsoft's Global Building Footprints~\cite{MSFTfootprints}. We note here that these building footprints have their own limitations. E.g., OpenStreetMap data may be outdated, and Microsoft footprints, retrieved from VHR optical imagery with a deep learning model, may suffer from missing buildings and geometric inaccuracies, particularly for small buildings. Considering the resolution of Sentinel-1 we only take into account the 15M buildings that exceed 50$\,$m\textsuperscript{2} in surface area, i.e., they are larger than half a pixel. \cref{app:footprints_limitations} illustrates some of these limitations.

\subsection*{Machine Learning Framework}

\begin{figure*}[!htbp]
    \centering
    \includegraphics[width=.9\textwidth]{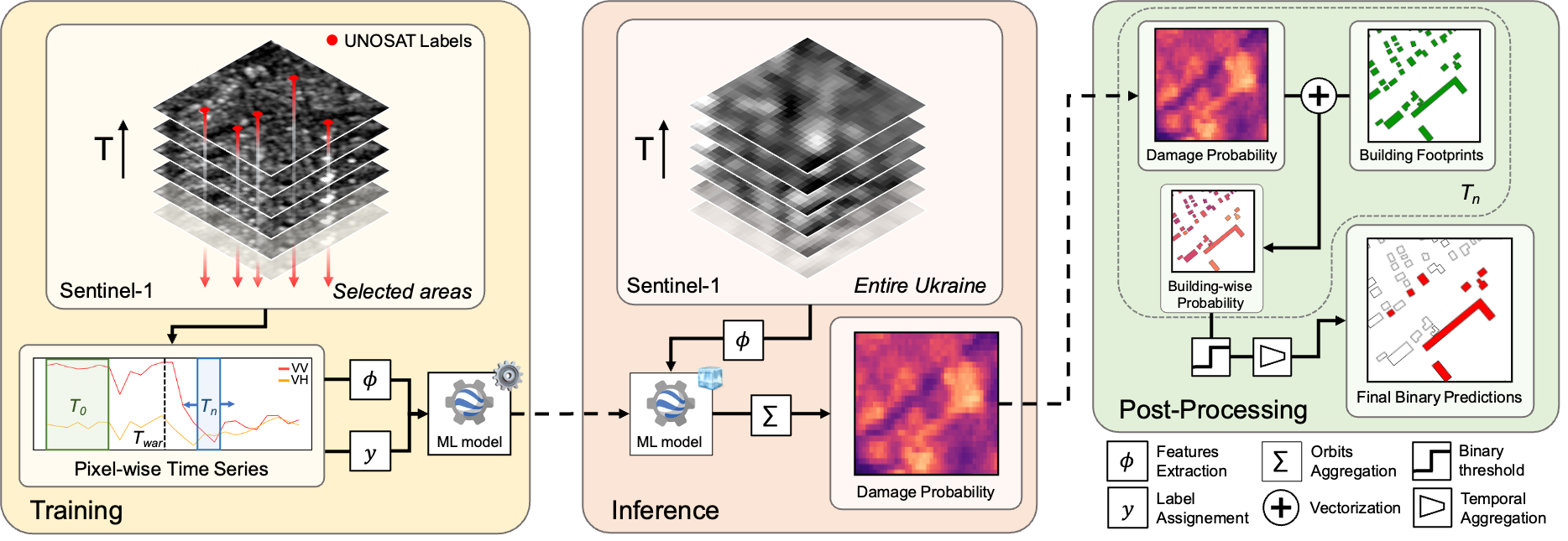}
    \caption{\textbf{Overview of the machine learning framework.} For training (left), we use per-pixel Sentinel-1 time series extracted at the location of UNOSAT point annotations. The model is fed with a pair of time series from the same location. The first one spans a fixed 12-month time interval $T_0$ from 2020, and the second one spans one of the 3-month time intervals $T_n$ between 2021 and 2023. Both time series are encoded with a custom features extractor, and damage labels are dynamically assigned according to $T_n$. At inference time (center), the model generates a damage probability heatmap valid at $T_n$ and spanning the entire country, aggregating the predictions of different Sentinel-1 orbits. The raw damage probabilities are intersected with building footprints. For the final map the estimates for different time intervals $T_n$ are thresholded and aggregated.}
    \label{fig:pipeline}
\end{figure*}

An overview of our machine learning framework is shown on \cref{fig:pipeline}. For a detailed ablation study of our key design choices, see \cref{app:ablation}.\\

\textbf{Definition of Time Intervals:} For clarity, we first define the temporal acquisition windows used in our study. Time interval $T_0$ covers one year from Feb.~24, 2020 until Feb.~23, 2021. The subsequent intervals $T_1$ to $T_{12}$ represent consecutive 3-month time windows, with $T_1$ ranging from Feb.~24, 2021 to May 23, 2021, and the last interval $T_{12}$ spanning Nov.~24, 2023 to Feb.~23, 2024. As such, intervals $T_1$ to $T_4$ represent the year preceding the invasion, while $T_5$ to $T_{12}$ represent the two years following it.\\

\textbf{Time Series Extraction:} For each location marked by UNOSAT, we stack the corresponding Sentinel-1 tiles and extract all backscatter values, for both the VV and VH polarizations. Limiting the training to the points annotated by UNOSAT corresponds to a fully supervised learning scheme where the target value is known for every training example. Relying only on pixel-wise time series discards all spatial context. We found that the time series signal carries most of the information relevant to our analysis, presumably because conflict-induced building damages are mostly small and localized. Per-pixel processing also scales particularly well on parallel processing architectures like the one underlying GEE.

Importantly, Sentinel-1 has the typical side-looking viewing geometry of SAR sensors, where the same location may look very different depending on the incidence angle and orbit direction (see \cref{app:backscatter_change} for examples). We therefore extract the backscatter values independently for each orbit, resulting in 2 to 4 time series per UNOSAT reference point. Overall, we obtain 33,304 distinct time series, denoted as $x_i^o$, where $i$ denotes the ID (respectively, location) and $o$ is the orbit index.\\

\textbf{Classification:} We formulate damage mapping as a supervised, binary classification. From each time series, we extract two segments: the fixed reference period $x_{i,\text{ref}}^o$ and the assessment period $x_{i,\text{new}}^o$. The task of the classifier is to estimate the probability that a war-related change has occurred in the time window between the two segments. For each segment and each band, we extract a set of $N$ statistical features and combine them into a final feature vector denoted as $\phi(x_i) \in \mathbb{R}^{4N}$ that forms the input to the classifier. This strategy has several advantages. First, by fixing the reference period before the start of the war one can ensure that it represents the state without any war-induced damages. Second, extracting fixed-length segments makes the method independent of the overall conflict duration. Third, using multiple different assessment periods greatly increases the number of training examples and also helps to make the model robust against seasonal variations.

As classification algorithm, we use a Random Forest as implemented in the SMILE library~\cite{Li2014Smile}, since that algorithm is available off-the-shelf in GEE, facilitating large-scale deployment. For every input feature vector the model outputs a score $\hat{y}_i^o \in[0,1]$, which can be interpreted as the likelihood that war-induced damage has happened between the associated reference and assessment periods. At inference time, we compute the overall damage probability map by averaging the estimates from different orbits, $\hat{y}_i = \frac{1}{N_o}\sum \hat{y}_i^o$.\\

\textbf{Training Details:} We perform all computations directly in GEE. We train the model on the four AOIs with the highest numbers of reference labels, Mariupol, NW Kyiv, Rubizhne, and Volnovakha. Together they represent 75.0\% of all UNOSAT annotations and 82.6\% of all backscatter time series. The remaining fourteen AOIs are reserved for evaluation, ensuring geographic diversity (see \cref{fig:unosat_data_overview}). We choose $T_0$ as a fixed reference period and randomly use periods from $T_1$ to $T_8$ (e.g., spanning one year before and after the beginning of the war) as assessment periods.

The label $y_i$ is dynamically assigned to each $x_i$ based on the end date $t_{max}$ of the assessment period:
\begin{equation}
    y_i = \begin{cases} 
    0 & \text{if } t_{max} \leq t_{\text{invasion}}, \\
    1 & \text{if } t_{max} > t_{\text{unosat}}, \\
    -1 & \text{otherwise} 
    \end{cases}
\end{equation}
where $t_{\text{invasion}}$ is 2022-02-24, the day the invasion was launched; and $t_{\text{unosat}}$ is the acquisition date of the post-event image used for annotation by UNOSAT. All time series with $y_i=-1$ are discarded because we cannot determine whether they contain the damage event. We transform $x_i$ into $\phi(x_i)$ using the following seven summary statistics: minimum, maximum, mean, median, standard deviation, kurtosis, and skewness. Finally, we configure the Random Forest to use 50 decision trees, with \textit{minLeafPopulation} set to 3 and \textit{maxNodes} set to 10,000. These values ensure sufficient robustness while remaining within the computational budget permitted by GEE.

\subsection*{Country-wide Inference}
\label{sec:infer_postprocessing}
After model training, we leverage the parallel computing capabilities of GEE to generate damage probability maps for the entire country. We compute one map for each period seen during training, as well as four additional maps for $T_9$ to $T_{12}$, the assessment periods between 2023 and February 2024. Each map, in EPSG:4326 projection and stored in UInt8 format, has an extent of 88,867$\times$201,284 pixels and a file size of 17.88GB.

\subsection*{Post-Processing}
To quantify the impact of the war on the Ukrainian building stock, we cross-reference our maps with the building footprints sourced from Overture Maps~\cite{OvertureMaps}. For each building $b_j$ and each period $T_n$, we assign a damage likelihood $\hat{y}_{j,T_n}$ by averaging the likelihood values of all pixels that fully or partially overlap the footprint, weighted by the overlap fraction:
\begin{equation}
    \hat{y}_{j,T_n} = \sum_i w_{ij}\hat{y}_{i,T_n}
\end{equation}
Here $\hat{y}_{i,T_n}$ represents the model output at pixel $i$ for the period $T_n$, and $w_{ij}$ is the proportion of that pixel that falls within $b_j$. By construction, this process also discards all damage estimates that fall outside of buildings, e.g.\ in agricultural areas or forests.

To obtain a final estimate of the number of buildings impacted over the first two years of the war, we aggregate the maps for the relevant assessment periods with the following rule:
\begin{equation}
    \hat{y}_j = \begin{cases} 
    1 & \text{if }\max(T_{[5,12]}) \geq t  \text{ and } \max(T_{[1,4]}) < t \\
    0 & \text{otherwise} 
    \end{cases}
\end{equation}

{\footnotesize
\section*{\normalsize Code Availability}
All the code necessary to reproduce our results can be found in the following repository: \url{https://github.com/prs-eth/ukraine-damage-mapping-tool/}. 

\section*{\normalsize Data Availability}
All data used in this study is publicly accessible. Links to all results, including damage heatmaps and building footprints with corresponding damage estimates, can be found in the repository hosting our code and/or in Zenodo (\url{https://zenodo.org/records/14811504/}).

\section*{\normalsize Author Contributions}
\textbf{Olivier Dietrich:} conceptualization (equal), data curation (lead), investigation (lead), methodology (lead), software (lead), visualization (lead), writing - original draft preparation (lead). \textbf{Torben Peters:} conceptualization (support), investigation (support), methodology (equal), software (support), supervision (equal), writing - original draft (equal). \textbf{Vivien Sainte Fare Garnot:} conceptualization (support), supervision (support), writing - review \& editing (support). \textbf{Valerie Sticher:} conceptualization (lead), data curation (support), funding acquisition (lead), project administration (equal), writing - original draft (equal). \textbf{Thao Ton-That Whelan:} funding acquisition (equal) project administration (equal), validation (support), writing - review \& editing (support). \textbf{Konrad Schindler:} conceptualization (equal), funding acquisition (equal), methodology (equal), project administration (equal), resources (lead), supervision (lead), writing - review \& editing (lead). \textbf{Jan Dirk Wegner:} conceptualization (lead), funding acquisition (lead), methodology (equal), project administration (equal), resources (support), supervision (lead), writing - review \& editing (lead).

\section*{\normalsize Competing Interests}
The authors declare no competing interests.

\section*{\normalsize Acknowledgments}
This work was supported in part by ETH4D's Engineering Humanitarian Action initiative through the Remote Monitoring of Armed Conflict project in collaboration with the ICRC, as well as by ESA's Discovery Programme under the Humanitarian Monitoring with Copernicus Satellites project (Contract No. 4000138343).

}

\onecolumn
\newpage
\printbibliography[title=References]

\end{refsection}

\onecolumn
\clearpage

\suppmaketitle

\begin{appendices}
\begin{refsection}[references_supp.bib]

\renewcommand{\appendixname}{Supplementary Note} 
\renewcommand{\thesection}{\arabic{section}} 
\renewcommand{\thefigure}{S\arabic{figure}} 
\renewcommand{\thetable}{S\arabic{table}} 

\clearpage

\section{UNOSAT Dataset}
\label{app:unosat_details}
Our reference dataset has been created by combining all existing UNOSAT datasets for Ukraine~\cite{unosat}. We combined them into 18 distinct areas of interest (AOIs). For our purposes, we only use the two strongest labels, \textit{destroyed} and \textit{severely damaged}. For the sake of completeness, we also report the third one here, \textit{moderately damaged}. For Mariupol and Kharkiv, for which multiple analyses were conducted by UNOSAT, we keep only the latest label assigned to a point, assuming that in the case of repeated assessments, the annotations were progressively refined. \cref{tab:AOIs} summarizes our AOIs and the number of labeled locations in each of them.

\begin{table}[ht]
    \centering
    \bgroup
    \def\arraystretch{1.75}
    \begin{tabular}{ccccccc}
        \toprule
        \thead[b]{AOI ID} & \thead{Locations} & \thead{UNOSAT\\product ID} & \thead{\# Destroyed} & \thead{\# Severely\\damaged} & \thead{\# Moderately\\damaged}  &\thead{Total} \\ \toprule
        UKR1 & \makecell{Mariupol\\Azovstal} & \makecell{3371, 3300\\ 3358} & 365 & 2184 & 3102 & 5651\\ \hline
        UKR2 & \makecell{Vorzel\\Hostomel\\Irpin\\Bucha\\Moschun} & \makecell{3356\\3359\\3360\\3363\\3417} & 647 & 1116 & 670 & 2433\\ \hline
        UKR3 & \makecell{Rubizhne\\Lysychansk\\Sievierodonetsk} & \makecell{3414\\3444\\3446} & 700 & 3182 & 1041 & 4923\\ \hline
        UKR4 & Volnovakha & 3415 & 263 & 541 & 66 & 870\\ \hline
        UKR5 & Avdiivka & 3443 & 36 & 505 & 86 & 627\\ \hline
        UKR6 & Chernihiv & 3354 & 266 & 374 & 256 & 896\\ \hline
        UKR7 & Kharkiv & \makecell{3357, 3455, 3454} & 56 & 168 & 241 & 465\\ \hline
        UKR8 & Borodyanka & 3355 & 43 & 55 & 58 & 156\\ \hline
        UKR9 & Makariv & 3403 & 18 & 64 & 21 & 103\\ \hline
        UKR10 & Mykolaiv & 3404 & 33 & 66 & 14 & 113\\ \hline
        UKR11 & Shchastia & 3405 & 4 & 12 & 1 & 17\\ \hline
        UKR12 & Sumy & 3406 & 3 & 6 & 10 & 19 \\ \hline
        UKR13 & Trostianets & 3407 & 12 & 29 & 6 & 47\\ \hline
        UKR14 & Kramatorsk & 3408 & 5 & 25 & 8 & 38\\ \hline
        UKR15 & Okhtyrka & 3413 & 10 & 20 & 9 & 39\\ \hline
        UKR16 & \makecell{Kherson\\Antonivka} & \makecell{3436\\3435} & 6 & 90 & 9 & 105\\ \hline
        UKR17 & Kremenchuk & 3437 & 1 & 7 & 3 & 11\\ \hline
        UKR18 & Melitopol & 3442 & 3 & 19 & 13 & 35\\ \hline
        \multicolumn{3}{c}{\thead{Total}} & 2471 & 8463 & 5614 & 16548\\ \bottomrule
    \end{tabular}
    \caption{\textbf{Overview of ground truth} Summary of the 18 distinct areas of interest (AOI) in Ukraine, merging data from 29 UNOSAT products. The table lists counts of destroyed, severely damaged, and moderately damaged labels per area, totaling 16,548 labeled damage instances, or 10,934 after excluding the moderately damaged ones.}
    \label{tab:AOIs}
    \egroup
\end{table}
\clearpage

\section{Detailed Comparison with PWTT}
\label{app:ttest}
The Pixel-Wise T-Test (PWTT), proposed in~\cite{Ballinger2024}, is a statistical approach designed to identify changes by comparing the mean values of two Sentinel-1 amplitude time series, adjusted by their standard deviation. We use it as a performance baseline since it utilizes the same data and has similar specifications, in particular, the scalability to large regions and the ease of deployment on GEE.

The PWTT can be calibrated to various conflict settings by optimizing a threshold value to fit existing damage assessments like our UNOSAT labels. For Ukraine in particular, the authors reported an optimal cutoff value of 1.63, based on data from eight different cities. However, their reported results cannot be directly compared to ours as they used different temporal ranges and different training examples for the \emph{undamaged} category: they assume that every building footprint without a damage label from UNOSAT is intact, disregarding the fact that damages could have occurred after the date of the UNOSAT analysis, or could have been overlooked.

For a meaningful comparison, we keep the recommended, optimal cutoff value but rerun PWTT using our time periods and our own, stricter definition of negative labels. We report the corresponding results in \cref{tab:ttest_full}. For readability, we repeat our results in \cref{tab:pixel-wise-eval-recopied}. It should be noted that Chernihiv and Kharkiv, which represent 61\% of our test set, were used by the original authors during the calibration of the PWTT. Nevertheless, our method consistently outperforms the baseline across all metrics.

\begin{table}[!htpb]
\centering
\small
\begin{tabular}{lcccccc}
\toprule
\thead{Label}     & \thead{Precision} & \thead{Recall} & \thead{F\textsubscript{1}} & \thead{Accuracy} & \thead{AUC}\\
\midrule
Damaged     & 61.9 & 76.3 & 68.4 & \multirow{2}{*}{75.5} & \multirow{2}{*}{75.7}\\
Undamaged   & 85.6  & 75.1 & 80 & \\
\bottomrule
\\
\end{tabular}
\caption{\textbf{Evaluation metrics for PWTT.} Quantitative performance of the PWTT approach on our test set, with the cutoff value of 1.63 as recommended for Ukraine. All metrics are given as percentages.}
\label{tab:ttest_full}
\end{table}

\begin{table}[!htpb]
\centering
\small
\begin{tabular}{lcccccc}
\toprule
\thead{Label}     & \thead{Precision} & \thead{Recall} & \thead{F\textsubscript{1}} & \thead{Accuracy} & \thead{AUC}\\
\midrule
Damaged     & 67.1 & 84.6 & 74.9 & \multirow{2}{*}{80.3 } & \multirow{2}{*}{81.3}\\
Undamaged   & 90.5 & 78.0 & 83.8 & \\
\bottomrule
\\
\end{tabular}
\caption{\textbf{Evaluation metrics for our approach.} Quantitative performance of our approach on our test set, with confidence threshold 0.5 (reprinted from \cref{tab:pixel-wise-eval}). All metrics are given as percentages.}
\label{tab:pixel-wise-eval-recopied}
\end{table}
\clearpage

\section{Choice of threshold}
\label{app:threshold}
\begin{figure}[htbp]
\centering
\includegraphics[width=.9\textwidth]{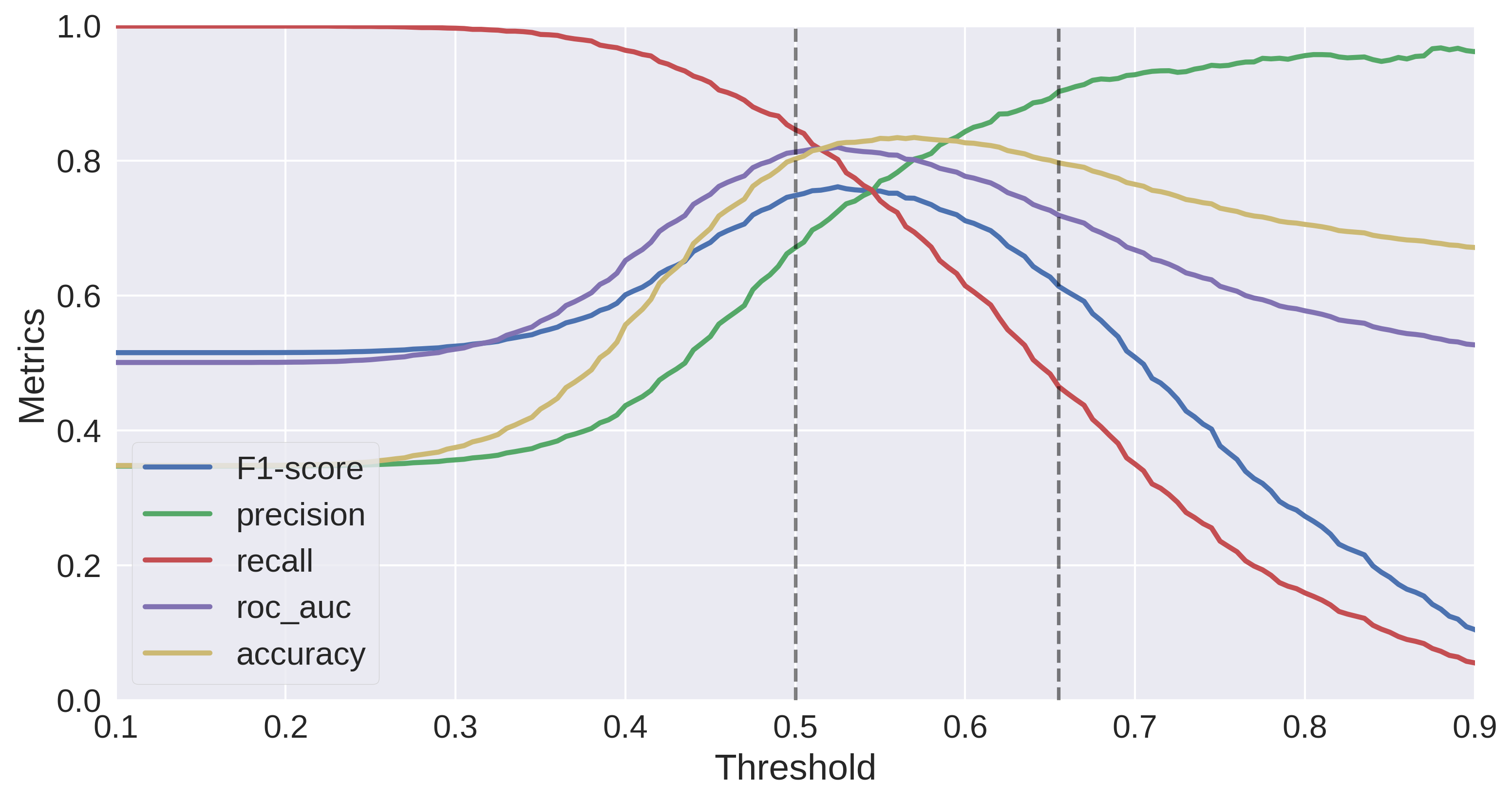}
\caption{\textbf{Performance of the model for different thresholds.} Pixel-level performance of our method on our test set, for different confidence thresholds. We set a target desired precision of 90\% and found a confidence threshold of 0.655. The two vertical lines indicate the thresholds 0.5 and 0.655.}
\label{fig:optim_thresh}
\end{figure}

There is an inherent trade-off between precision and recall, and the optimal threshold depends on the specific application. To keep false alarms at a low enough rate, we set a target precision of 90\%, for which we find an optimal confidence threshold of 0.655. That value has been used to generate all metrics in the present paper unless stated otherwise. With this threshold, the user can expect a recall of 46.5\%. In practice, users of our tool can dynamically adjust the confidence threshold to suit their specific needs. For instance, one could start with a high threshold to discover hotpots with a very high likelihood of destruction, then gradually decrease the threshold to increase the recall and better assess the extent of destruction. \cref{fig:optim_thresh} illustrates how our metrics vary with different thresholds.

\clearpage

\section{Temporal Evolution of Damage}
\label{app:temporal_evolution}
\begin{figure}[htbp]
\centering
\includegraphics[width=.9\textwidth]{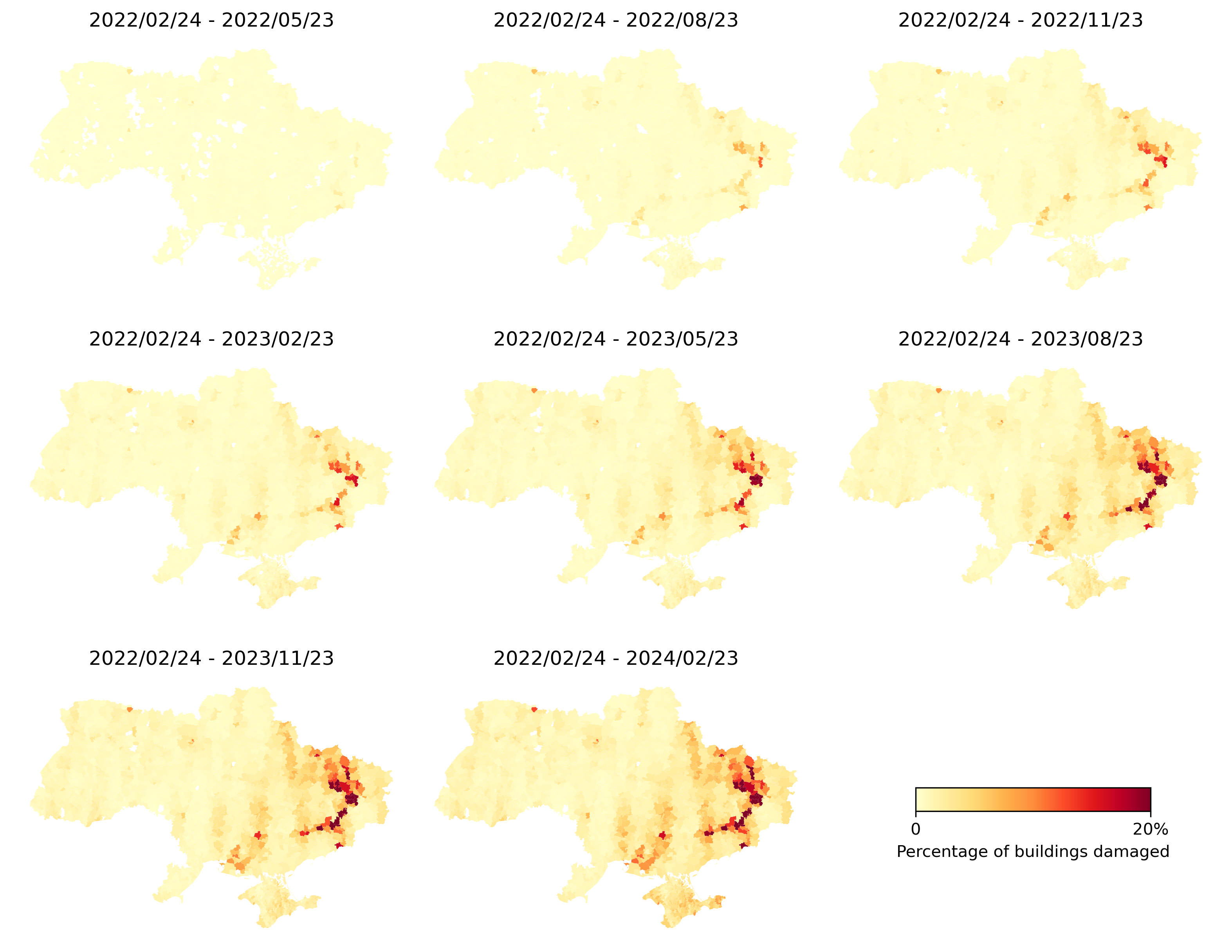}
\caption{\textbf{Evolution of damage distribution in Ukraine over time.} Percentage of buildings likely damaged for each period, cumulated since the beginning of the war and aggregated by \textit{hromadas}. The predictions were thresholded at 0.655, and only buildings larger than 50$\,$m\textsuperscript{2} were considered.}
\label{fig:temporal}
\end{figure}

The series of maps in Figure \cref{fig:temporal} illustrates the progression of building damage over the first two years of the conflict. The maps clearly delineate the primary locations of battles and the frontline, highlighting the areas most affected as the conflict persists.

\clearpage

\section{Limitations of Building Footprints}
\label{app:footprints_limitations}

\begin{figure}[htbp]
    \centering
    \includegraphics[width=.9\textwidth]{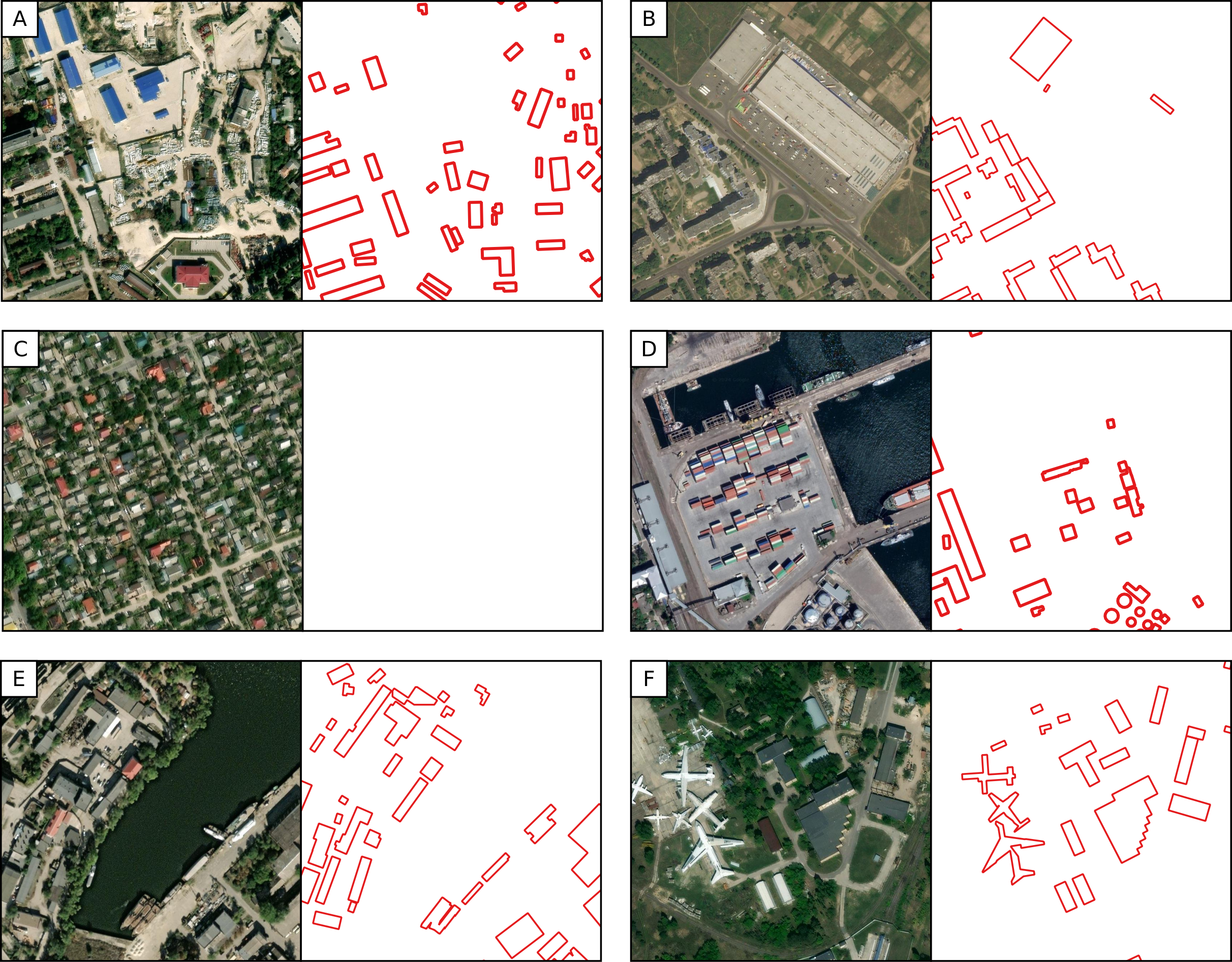}
    \caption{\textbf{Quality issues of existing building footprints.} All images cover 500$\times$500$\,$m\textsuperscript{2} and are displayed alongside the corresponding building footprints sourced from Overture Maps. Only buildings larger than 50$\,$m\textsuperscript{2} are depicted. (A) Outdated buildings in Sevastopol. (B) Absent shopping mall near Chernihiv. (C) Missing footprints in Kherson. (D) Containers in Odessa. (E) Boats in Zaporizhzhia. (F) Antonov International Airport, Hostomel. Satellite images were obtained from ESRI/Maxar Technologies (A and C), Bing/Maxar Technologies (B, E, and F), and Google/Maxar Technologies (D).}
    \label{fig:footprints_limitations}
\end{figure}

The building footprints provided by Overture Maps~\cite{OvertureMaps}, which for Ukraine were compiled from OpenStreetMap~\cite{OpenStreetMap} and Microsoft~\cite{MSFTfootprints}, inevitably come with limitations. For instance, OpenStreetMap data may be obsolete due to the demolition of older structures (\cref{fig:footprints_limitations}.A) or the construction of new ones (\cref{fig:footprints_limitations}.B). While Microsoft footprints serve as a valuable complement to OpenStreetMap data, they were automatically retrieved with a deep neural network from Bing Maps imagery, which may not always offer recent images, and suffers from gaps due to cloudy or outdated acquisitions, for example in Kherson (\cref{fig:footprints_limitations}.C). Moreover, it is not uncommon for large non-building structures, e.g.\ containers (\cref{fig:footprints_limitations}.D), boats (\cref{fig:footprints_limitations}.E) or even aircraft (\cref{fig:footprints_limitations}.F), to be wrongly identified as buildings.

\clearpage

\section{World Bank Estimates}
\label{app:world_bank_comp}
In this section, we compare our estimates, with a total of $\approx$400,000 buildings damaged between Feb 2022 to Feb 2024, to figures provided by the World Bank (WB) in their third Rapid Damage and Needs Assessment report~\cite[p.~74-80]{WorldBank2023}.

According to the WB report, a total of 2,106,920 housing units were damaged by December 2023. Among these, 547,010 were destroyed, 679,382 sustained medium damage and 880,528 suffered minor damage. Importantly, the WB figures refer to housing units (HU), not buildings. A single-family home counts as one HU, but multi-family buildings may contain dozens of HU, making a direct comparison to our building-based estimates difficult.

In addition to damage estimates, the report provides the geographical distribution of damage cost. The WB attributes over 75\% of the total damage cost to the following oblasts: Donetska (30 $\%$), Kharkivska (27 $\%$), Luhanska (12 $\%$), and Kyivska (8 $\%$). Another 18 $\%$ is distributed among Mykolaivska (4 $\%$), Chernihivska (4 $\%$), Khersonska (4 $\%$), Zaporizka (3 $\%$), and Dnipropetrovska (3 $\%$). We use this spatial distribution of damage cost as a proxy to compare with our own estimates.

\begin{figure}[!htpb]
    \centering
    \includegraphics[width=.75\linewidth]{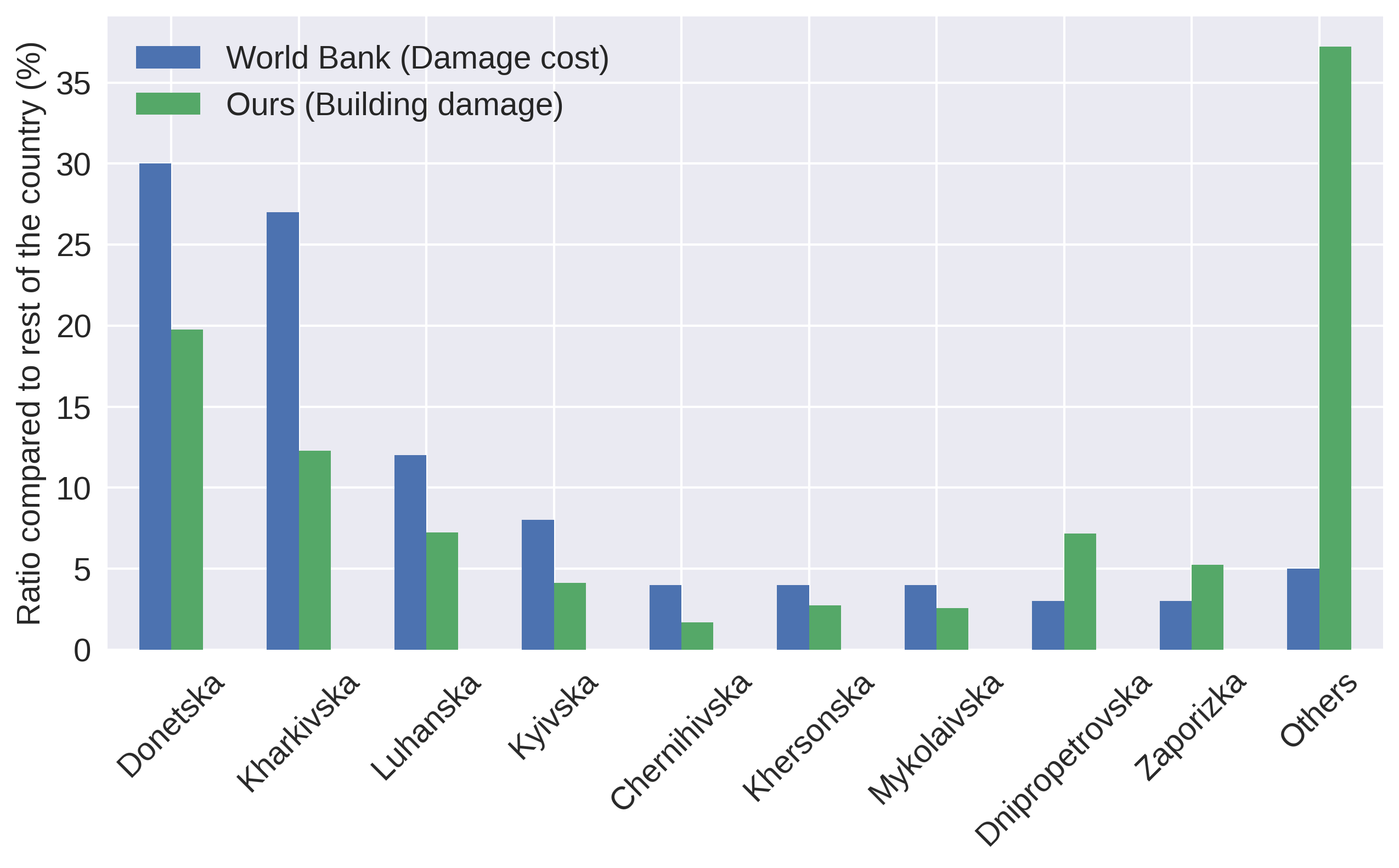}
    \caption{\textbf{World Bank estimates vs Ours.} Comparison of the geographical distribution of the World Bank estimates against ours. The World Bank provides spatial distribution for their damage cost estimates, while our spatial distribution shows buildings likely damaged.}
    \label{fig:world_bank_vs_ours}
\end{figure}

\cref{fig:world_bank_vs_ours} shows the differences between our estimates and those of the WB. Excluding the \textit{other} oblasts outside the direct conflict zone, both distributions align quite well, with indeed the majority of damage concentrated in Donetska and Kharkivska. However, our model estimates considerably more damage than the WB for the \textit{other} oblasts, i.e., and the ones further from the frontline. It appears that construction and other non-conflict-related changes dominate in those regions, which our model is unable to distinguish from building damages. This highlights that there is a trade-off between, on the one hand, comprehensive screening to detect also unexpected conflict-induced destruction far from the known frontline; and, on the other hand, accurate assessment by ruling out areas that are a-priori unlikely to be affected.

\clearpage

\section{Changes in Backscatter due to Destruction}
\label{app:backscatter_change}

The backscatter signal from a building is influenced by numerous variables, including building characteristics (e.g., geometry, orientation, surface texture, material composition), satellite parameters (e.g., frequency, polarization, incidence angle, orbit direction), and environmental factors (e.g., vegetation, soil moisture, snow, or heavy rainfall). Intuitively, one might expect that intact structures yield stronger specular returns, with flat surfaces and right-angle corners acting as corner reflectors, whereas structural damage would disrupt these patterns, producing a more diffuse signal. However, several studies indicate that damage signatures in SAR data are highly variable and depend on the type and extent of destruction~\cite{Plank2014, Ge2020}. Damaged buildings can even exhibit backscatter signals as high as undamaged structures, or even higher~\cite{Brunner2010}.

\begin{figure}[h!]
    \centering
    \includegraphics[width=\textwidth]{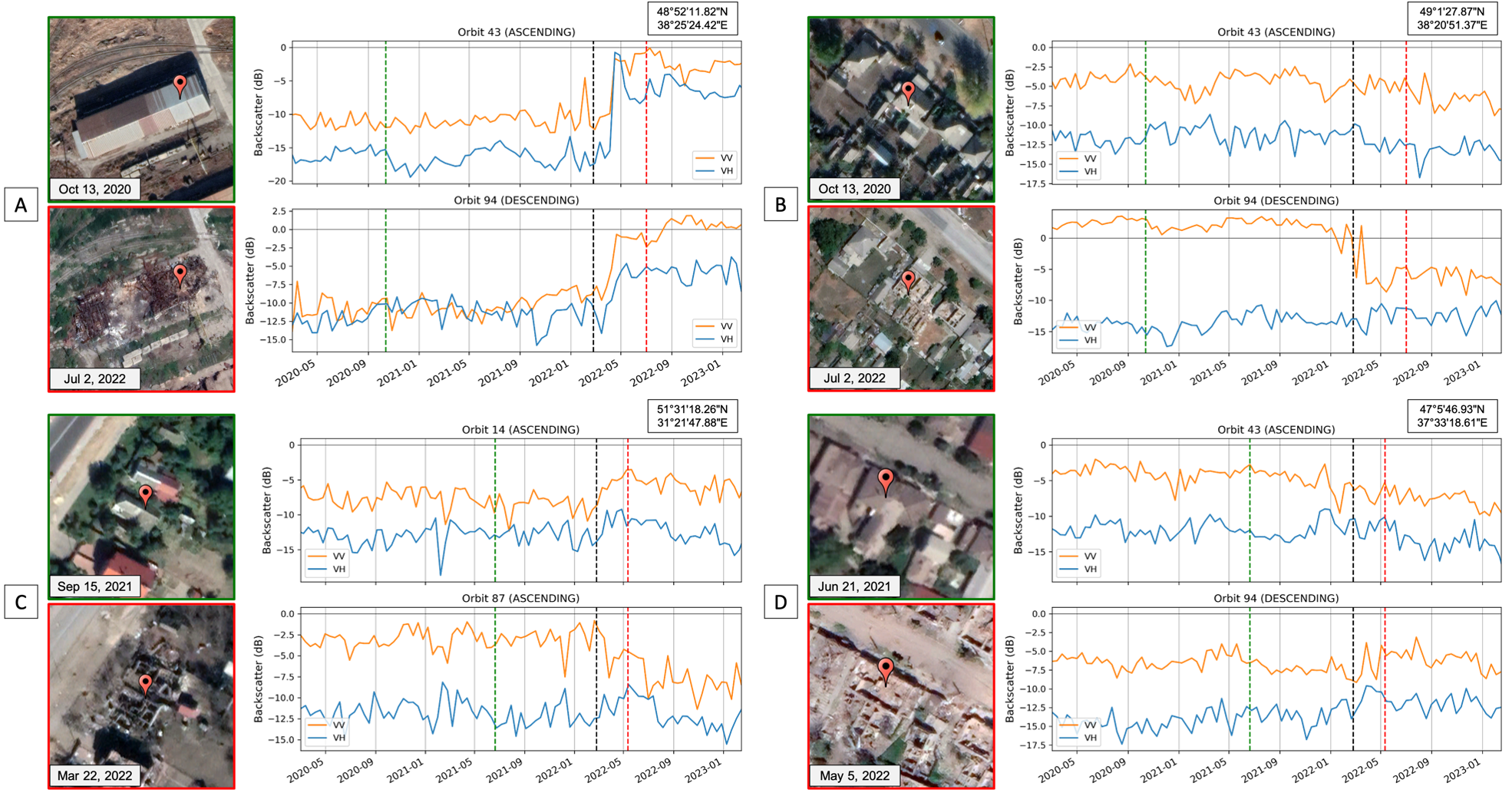}
    \caption{\textbf{Example of Sentinel-1 time series} Sentinel-1 backscatter time series at selected UNOSAT locations for VV and VH polarizations, with VHR satellite images from Google Earth showing pre- and post-destruction conditions. Vertical dashed lines indicate the date of Russia's full-scale invasion (black) and the dates of the displayed pre-destruction (green) and post-destruction (red) images. For readability, only two orbits per location are shown, even if in some locations up to four orbits are available. Amplitude can decrease (B, C), increase (A, C) or remain relatively stable (B, D) after destruction. The same event may manifest differently depending on the flight direction (B), or even on the incidence angle within the same flight direction (C). Basemaps from Google Earth/Maxar Technologies.}
    \label{fig:backscatter_examples}
\end{figure}

Our experiments corroborate these findings. \cref{fig:backscatter_examples} presents pixel-wise backscatter time series that illustrate how conflict-related destruction appears in the data. High-resolution pre- and post-destruction satellite imagery from Google Earth is provided for reference. We observe that the backscatter amplitude following destruction can vary widely -- increasing, decreasing, or remaining largely stable -- depending on the location, the satellite orbit, and the polarization. These examples emphasize the need to utilize all available orbits in order to maximize reliability.

\clearpage

\section{Ablation studies}
\label{app:ablation}

We have performed an ablation study to support our design choices and understand the impact of various settings on our model. We kept the same train/test split and label definitions as described in  \nameref{sec:eval}. To reduce computational cost, we opted for a pixel-wise comparison to the UNOSAT labels, contrary to \nameref{sec:eval}, where we ran the inference densly over the entire area of interest and allowed for shifts up to $\pm$1 pixel in $x$ and $y$ to account for geo-location uncertainty of the labels.

\begin{figure}[h!]
    \centering
    \includegraphics[width=.98\textwidth]{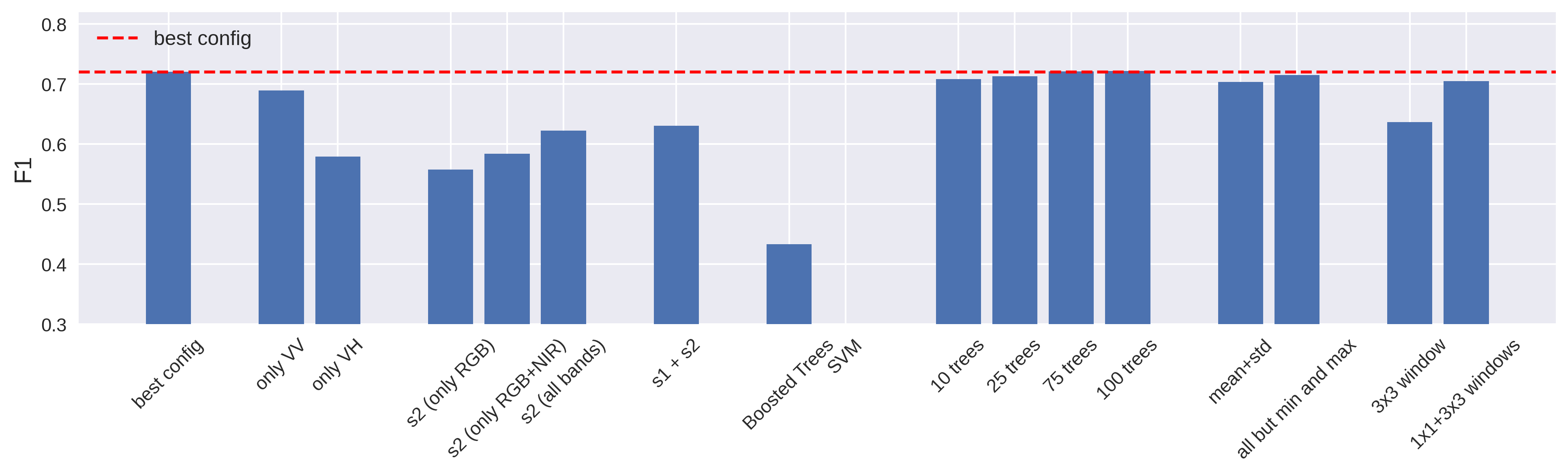}
    \caption{\textbf{Results of the ablation studies.} $F_1$-score for different configurations. The dashed red line indicate the best configuration.}
    \label{fig:ablation}
\end{figure}

The $F_1$-score of our optimal configuration, which utilizes both VV and VH Sentinel-1 bands, all seven features, 50 decision trees and extracts the time series pixel-wise, is 0.72. \cref{fig:ablation} summarized the outcome for different settings. We observe the following characteristics:

\textbf{Input bands:} Our tests confirmed that VV polarization provides more relevant information for detecting destruction compared to VH. However, using both bands together produced the best results, indicating that they complement each other to some degree.

\textbf{Input data:} We tried replacing Sentinel-1 SAR data with Sentinel-2 optical images~\cite{sentinel_2}. Pixel-wise Sentinel-2 time series were extracted from the Level-2A collection available on GEE~\cite{s2Harmonized}, from which we filter out cloudy pixels using GEE's CloudScore+ mask~\cite{Pasquarella2023}. Various band combinations were tried, but incorporating Sentinel-2 data never enhanced model performance. This was expected, as our pipeline was design for Sentinel-1 properties, such as stable pre-event signals and a focus on temporal context over spatial context. Intuitively, it is harder to distinguish intact buildings from damaged ones by looking at the brightness change at an individual pixel location. Also combining Sentinel-1 and Sentinel-2 led to poorer results than using Sentinel-1 alone, likely due to the confounding effect of a six-fold increase in features, most of which contain little information for the task. 

\textbf{Model:} We evaluated alternative classifiers available in GEE, i.e., Boosted Trees~\cite{Li2014Smile} and SVM~\cite{libvsm}. Boosted trees showed weaker performance, while taking approximately 3$\times$ longer to run. We were unable to obtain results for the SVM: when applied to our data it quickly exceeded the computational budget allowed by GEE.

\textbf{Number of trees:} We tested models with 10, 25, 50, 75, and 100 trees. Performance gains were minimal beyond 50 trees, so we chose 50 as the optimal number, balancing accuracy and computational cost. Above 100 trees, we again hit GEE's bound on the compute budget.

\textbf{Features:} We tried different configurations, from a simple combination of the mean and standard deviation to the entire set of seven features. We observed consistent gain as the number of features increased.

\textbf{Extraction window:} Instead of extracting the time series pixel-wise, we experimented with taking the spatial mean over a 3$\times$3 neighborhood to suppress noise. The resulting heatmaps are smoother, but this comes at the cost of missing many small-scale and sparse damage events.

\newpage
\printbibliography[title={Supplementary References}]

\end{refsection}
\end{appendices}
\end{document}